\documentclass[12pt]{article}
\usepackage{amsmath, amsthm}
\usepackage{graphicx}
\usepackage{comment}
\usepackage{listings}
\usepackage{authblk} 

\newcommand\norainprob{87.64}
\newcommand{\BigO}[1]{\ensuremath{\operatorname{O}\bigl(#1\bigr)}}

\begin{document}
\title{%
How Much Did it Rain? \\
\large \textit{Predicting Real Rainfall Totals Based on Radar Data} \\
\large CS289 Project Report}
\date{May 13, 2015}
\author{Adam Lesnikowski%
  \thanks{Email: \texttt{adam@math.berkeley.edu}, \texttt{adam.lesnikowski@gmail.com}}}
\affil{University of California, Berkeley}

\maketitle

\begin{abstract}
	We applied a variety of parametric and non-parametric machine learning models to predict the probability distribution of rainfall based on 1M training examples over a single year across several U.S. states. Our top performing model based on a squared loss objective was a cross-validated parametric k-nearest-neighbor predictor that took about six days to compute, and was competitive in a world-wide competition. 
\end{abstract}
	
\section{Motivation}
    Today water is an essential commodity, for growing food, providing drinking water, and 
	countless other uses. As climate change progresses, past historical records of 
	rainfall totals may not be altogether great estimates of future rainfall total.
	So how can we tell, how much rain actually fell?
	One option is direct measurement of rainfall total, the thing we really care about, via rain gauges.
	However these are expensive and limited in time and location.
	An alternative source is information from radar arrays, which is cheap and abundant.
	But currently this alternate approach suffers a big drawback, in that
	the existing algorithms to infer actual rainfall total from radar data do not perform that well.
	As an example, some current algorithms actually predict negative rainfall totals some the time, 
	which is clearly an impossible total accumulated rainfall prediction.
	Furthermore for some downstream agricultural and climate models, having a probabilistic
	density estimate is desired, instead of the point prediction the currently used algorithms provide.
	Hence there is a need to squeeze as much useful data from the clouds of data as possible, meteorological pun
	intended. The machine learning problem, then, is given radar measurements for a location and time, 
	such as radar reflectivity, precipitation type, and average precipitation shape, predict a probability density of
	how much rain fell. This is the challenge that we address here.  
	
\section{Problem Setup}
    \subsection{The Data Set}
	    The data set consists of 1,126,694 training points.
		Each data point is a collection of numerical radar features collected in one hour for some particular location.
		These locations vary amongst several midwestern states in the United States, 
		and were taken between April and November 2013.
		The data set provides no location or time features, and were shuffled, so that there is no immediate
		way to recover location or time features. 
		
		There are 19 provided features, with three of these features being rain rates
		predicted from three current algorithms.
		These three past algorithm features, RR1, RR2, and RR3, are respectively, the
		`HCA-based', `Zdr-based', and `Kdp-based' algorithms.
		The other 16 features are given as time series numerical data.
	    An example data point could have its `TimeToEnd' feature s
		`58.0 55.0 52.0 49.0 41.0,' indicating radar information 
		taken at 58, 55, $\dots$, 41 minutes from the end of the hour.
		For this same row, the features 
		`Reflectivity' as `0.0, 0.0, 1.2, 4.5, 0.0' and
		`RR1' as `0.0. 0.0, 2.2, 0.3, 0.0' mean these measurements taken 
		at the time points in the `TimeToEnd' series.
		The label for each row is one float number, the amount in mm of 
		rain collected for that hour. 
		
		The test set consists of 630,521 points.
		The test set draws from the same collection of radars covering the same region,
		but in the next year, 2014.
		It is not indicated whether the test points were drawn 
		according to the same time or location distribution as in the training set for 2013.
		For the online Kaggle competition, submissions are predictions of the probabilistic
		distribution of the hourly rain total. 
		Each row of the submission is a list of values $P(y \leq Y)$,
		for $Y$ integer values $0, 1, 2, \dots 69$, and $y$ the rainfall total, in mm.

	\subsection{The Evaluation Function}
	    Each submission matrix $A$ with rows consisting of entries $Pr(y\leq 0)$, $Pr(y\leq 1)$, $\dots$, $Pr(y\leq 69)$ is evaluated according to the function
		
		$$\textit{Score}(A) = \sum_{i}^{m} \sum_{j=0}^{n-1}  (Pr(y_i \leq j) - H(j - y_i))^2,$$
		
		for $m$ the number of test points, $n$ the number of bins, in this case 70, 
		$Pr(y_i \leq)$ our predicted probability, $y_i$ the true label for the $i$th row,
		and $H(x)$ the Heavyside step function, which is 1 for $x \geq 0$ and 0 otherwise.
		
		The score can be thought of as the squared loss, averaged across all the data points and bins, 
	    either from 0 if the bin label is less than the true label,
		or 1 if the true label is less than the bin label. 
		For instance a perfect prediction for the true label of $2.5$
		would be the row $0, 0, 0, 1, 1, \dots, 1$
		corresponding to $Pr(y \leq 0) = Pr(y \leq 1) = Pr(y \leq 2) = 0$ and
		$Pr(y \leq 3) =  \dots = Pr(y \leq 69) = 1$.
		As we are seeking to minimize a squared loss function,
		and will not in general have complete certainty in our predicted label,
		so we will `hedge our bets' with each prediction,
		and predict according to our estimated probabilities of the test point landing in each of the seventy bins.
		
	\subsection{Model Selection for this Problem}
	    The relatively large amount of training points, about 1 million, and the low amounts of provided features,
		about 30, put us in a situation where $m >> n$, for $m$ the number of training samples, and 
		$n$ the number of features.
		This suggests that overfitting should be less an issue with this data set 
		than with smaller data sets.
		For parametric models, if we adopt the rough rule of thumb that 15-30 data points are enough to train
		a model with acceptable variance, ~1,000,000 data points should be enough to train models with roughly ~33,000 - 66,000 parameters. We note that this rule of thumb is only a heuristic, and we 
		explored the effect of training set size on our parametric model to see if this would lead to improved performance whenever possible.
		Given the somewhat large amount of training points and test points, 
		classifiers with algorithms that scale well were preferred in exploring classifiers. 
		Unlike problems in computational vision or audio, feature selection does not seem
		to be a major issue with this data set, hence classifiers such as neural networks were not explored. 
		For our $k$-n neighbor approach, as the number of features is relatively small, 
		while the number of training and test samples is relatively large,
		we prebuilt a $k-d$ tree to aid to make the problem computationally efficient. 
				
	\subsection{Dealing With Missing Data}
	    One major issue with this data set, and a vexing one in general is,
		how to deal with missing data.
		Should we ignore these data points, or to fill them in in some coherent way?
		For instance, discarding all data points with some missing value leaves
		only ~200K of the original 1.1M data points,
		with the number of missing values per column ranging from 0 to ~900K.
		The worst offending columns in this regard were the two
		past algorithm prediction columns RR2 and RR3.
		Our approach, unless otherwise noted, was to discard these two columns,
		since we are seeking to improve upon the existing predictors.
		Then, again unless otherwise noted, we filled in the rest of the missing column values with zeros.
		Though heavy-handed, this lead to competitive classifier performance.
		More complicated schemes of inferring missing data,
		from column averages through building separate predictors
		for missing data, were not pursued, though appear as promising ways to improve performance. 

\section{Simple Benchmarks}
    \subsection{`No Rain', The Null Hypothesis}
	    Inspection of a histogram of the training set labels gives us that 
		about \norainprob\% of the time, the rain label was 0.0 mm.
		This makes intuitive sense, in that most of the time, for most regions of the United States,
		it is not raining, under the assumption that the training samples were not favorably drawn
		from rainy periods or locations.
		Hence we tried a `No Rain' null-hypothesis prediction on the test data
		consisting of all 1 predictions, i.e.~the predicted probability 
		that it rained less than any amount was always 1.
		Somewhat surprisingly this classifier was competitive.
		With a score of 0.01017651, this submission ranks tied between places 207 through 220 
		on the online leaderboard, among 331 total entries, as of May 10, 2015.
		Clearly this prediction is only a first-order approximation, but it provides a useful benchmark, 
		and illustrates the point taking a common-sense look at the data set and simple statistics about it
		can provide simple models that outperform more complicated ones.
		
	\subsection{`Sigmoid', Provided benchmark}
	    A sample solution, which we dubbed `Sigmoid,' 
		was provided by the competition administrators, with strategy as follows.
		Consider the RR1 values of each test point, which were the rainfall rates predicted by the first past algorithm,
		average these, normalize by the time period covered by the radar, and use this
		point estimate to generate a density function using a sigmoid function.
		This solution scores 0.01177621, good for 252 out of 331 entries.
		This was worse than the `No Rain' submission, and we think this may be due to some combination of the following
		factors: a) normalizing by the time period of radar coverage instead of the whole 60 minutes
		overestimates the hourly rainfall total, since the time series under `RR1' gives hourly rainfall estimates, 
		b) the sigmoid function is a c.d.f.~of a tail-heavy normal-like 
		distribution, and may not model the real distribution of rainfall averages that well, and c) no
		estimate of the variance of the data was considered, so the parameter of the sigmoid function
		was constant for each test point, and hence not sufficiently tuned. 
		Further improvements to this classifier were not pursued. 
		    
\section{The Histogram Approach}
    \subsection{ `Histogram', Sample Set Histogram}
	    The next approach tried was `Histogram' strategy.
		This was to compute the proportion of the training set in each of the bins,
		and use this as the prediction for each test sample, regardless of the feature given.
		In this case we have a parametric model with
		a naive parameter estimation algorithm linear in $N$, the number of training points, 
        and $\BigO{1}$ complexity for making a prediction on a test sample, which is computationally excellent. 
		`Histogram' did even better than the two previous solutions,
		scoring 0.00971225, which was good for 184 out of 331.
		Again this emphasized the value of thinking about the problem, 
		in particular the squared-loss in the score function,
		for making cheaply computed, yet competitive classifiers.
		
    \subsection{Inferring Test Histogram}
	    As has been noticed for this and other Kaggle competitions, it is possible to use a submission
		scores to infer non-trivial information about the test sample.
		For instance, knowing the returned score of the all 1's `No Rain' submission
		above, submitting a solution with all 0's for the first column and all 1's elsewhere,
		and comparing these two scores, it is possible to determine the 
		proportion of test samples in bin 0.
		Likewise given 70 total samples, it would be possible to infer the
	    entire test sample histogram.
		Furthermore since for this competition, the scores returned for a submission are based on about 70\% 
		of the test data, these estimates would be quite good.
		One iteration of this approach gave an estimated 85.67\% of test data in bin 0,
		while 87.64\% of the training set were in bin 0.
		This difference is interesting in its own right in that it might indicate
		real differences between the 2013 training and 2014 test data,
		and place an upper bound on the accuracy of any machine learning algorithm that
		learns solely on the training data.
		Further querying the Kaggle website for test set histogram estimates though 
		were not further pursued in favor of creating an
		algorithm that actually learned from the training data.
	
\section{Ensemble of Previous Algorithms}
    \subsection{`Simple Average'}
	     Since each test sample came equipped with three features RR1, RR2, RR3,
		 the rain rate estimates given by the past algorithms, we tried
		 predictions based on an ensemble of these past algorithms.
		 We first tried `Simple Average,' which preprocessed the data by computing
		 the arithmetical mean of the time series of predictions of each algorithm,
		 and then averaged the three algorithms' predictions with equal voting weights.
		 Inspection of the predictions revealed some negative values,
		 so that this simple average was floored at zero.
		 No statistical estimates of bin probabilities were obtained,
		 so the point prediction was transformed into a 
		 step probability distribution by predicting $Pr(y \leq Y) = 0$
		 for $Y$ less than our estimate at $1$ otherwise. 
		 The submission based on this approach scored 0.01080682, which places 224th. 
		 This approach did somewhat better than `Sigmoid,' but also somewhat worse than either `No Rain' or `Histogram.'
		 
	\subsection{`Optimal Voting'}
	    An improvement on `Simple Average' is to give the rain rate estimates 
		given by the past algorithms differential voting weights.
		Hence we constructed `Optimal Voting,' based on solving the following optimization problem: 
		given the training set labels $y$, features $R$ RR1, RR2, RR3,
		what are the voting weights $w$ that minimize the distance
		between $w$ and $y$? Taking the distance to be the $l_2$ norm, this is equivalent to the problem:
		
		$$ \min_w ||Rw - y||^2_2,$$
		
		for $R$ the matrix of past algorithm predictions.
		Since least squares can be overly sensitive to outliers, 
		we pre-processed the data by throwing out outliers.
		In our case, we used the world record rainfall rate over one hour as an
		a priori bound, which as of 2015 is 305mm in one hour, recorded in Holt, Missouri in 1947. 
		These outliers were suspected to be gauge errors.
		The coefficient obtained for RR1 was larger than the other coefficients by roughly three order of magnitude,
		and were insensitive to retraining based on randomly selecting subsets of the data.
		This is some evidence that RR1 is the best predictor of the three past algorithms.
        `Optimal Voting' scored 0.01014877, good enough for 206th, 
		 beating both `Sigmoid' and `Sample Avg' by good margins,
		`No Rain' by a small margin, but doing worse than `Histogram'.
		We note that an immediate extension to these two ensemble approaches is to get good statistical
		estimates of the probabilistic distribution of the rainfall total,
		and not just using step distribution based on a point prediction, though this
		was not pursued in this report.

\section{Multi-Class Logistic Regression Predictors: A Parametric Approach}
    \subsection{The Setup}
	    To obtain a probability distribution estimates,
		we fitted parametric models to the data,
		in this case multinomial logistic regression models.
		Logistic regression models take the log odds of each class
		relative to some reference class
		as a linear function of the feature vector, 
		so that these are a particular kind of generalized linear models.
		In the multinomial setting, once parameters $\theta_i$ are estimated,  
		the probability class $j$ among $k$ class is given as
		
		$$ \frac{exp(\theta_j x)}{1 + \sum_i^k exp(\theta_i x)},$$
		
		where each feature vector is bundled with a constant number to allow fitting
		a constant bias, and normalized by the denominator to give some probability estimate
		between 0 and 1.
		The reference class, whose choice does not matter, has probability prediction
		given by replacing the above fraction by 1.  
		The parameters are fit by maximal likelihood estimation, 
		which amount to solving a numerical optimization problem by, for instance, 
		the Newton-Raphson method.
		Fitting for $k$ classes and $n$ features amounts to finding
		estimates for roughly $k \cdot n$ parameters. 	
    \subsection{Challenges}
       We did not obtain promising results from fitting our logistic regression. Results from our logistic regression model were worse than simple benchmarks. Increasing training set size did not seem to help much, which seemed to use like a sample size saturation effect. Adding $L1$ regularization did not help much either. Our diagnosis is that this logistic regression model appeared to be predicting equal probabilities for the first few bins, so it suffered a big score hit due to this. Proposed possible fixes include try other kinds of regularization, another numerical solver, or randomizing initial starting points in MLE search, though we did not have the resources to pursue these additional avenues.

\section{Nearest Neighbors Classifiers: A Non-Parametric Approach}
    \subsection{The Setup}
	   The $k$-nearest neighbors algorithm, on a high level, finds the $k$ nearest points to a test vector,
	   and assigns the test vector as some function of its neighbors' labels.
	   In our case, we estimate the probability distribution of the test point label into the 70 bins
	   by tallying the proportion of the neighbor's labels that fall into each bin.
	   Hence this is a kind of `local' extension of the histogram approach above,
	   where we are taking histograms of only the test point's nearest neighbors.
	   Given that our test sample is about 600K points, and we have a relatively small number of features,
	   we first built a $k$-d tree based on the training data, 
	   which then allows us to perform much speedier test sample predictions.
	\subsection{Finding Optimal K's}
\begin{figure}
	        \includegraphics[width=0.95\textwidth]{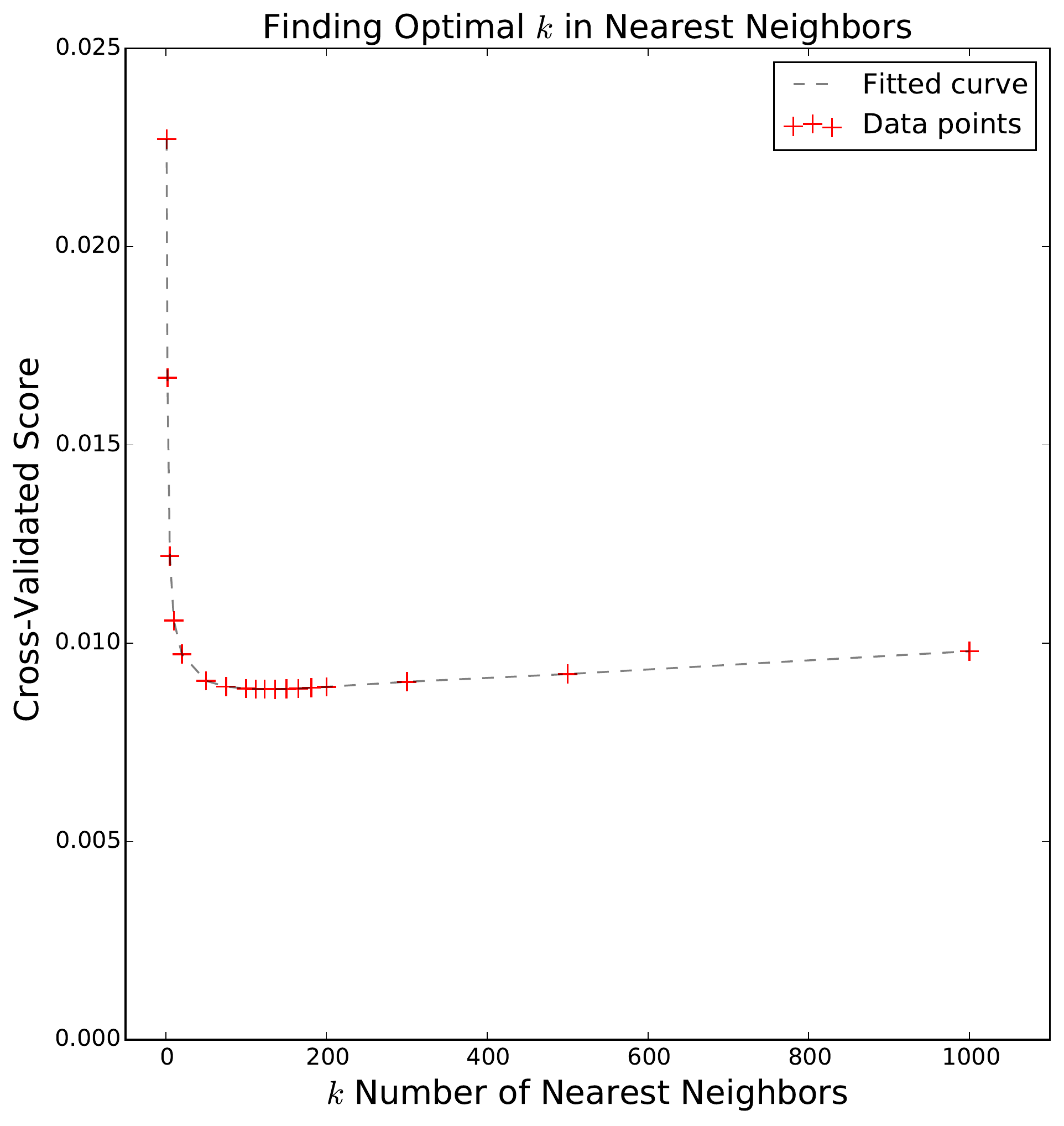}
			\caption{Finding Optimal $k$ in Nearest Neighbors}\label{fig:scoresVsK}
		\end{figure}

	    To find an optimal $k$ value, we randomly split off 
		a training set of 33,000 points and a validation set of 100,000 points from the training data.
		This asymmetry of sizes was a rough guess given the relative  
		costs associated with $k$-d tree construction versus test sample evaluation. 
		Iterative searches for $k$ values produced the following figure of a $U$ shaped performance curve.
	    Convolutions windows of around 150 performed best,
		with smaller values in general not providing enough context, and large values washing out too many details.
		See Figure~\ref{fig:scoresVsK}.
		
	\subsection{Effect of Training Set Size}
\begin{figure}
	       \includegraphics[width=0.95\textwidth]{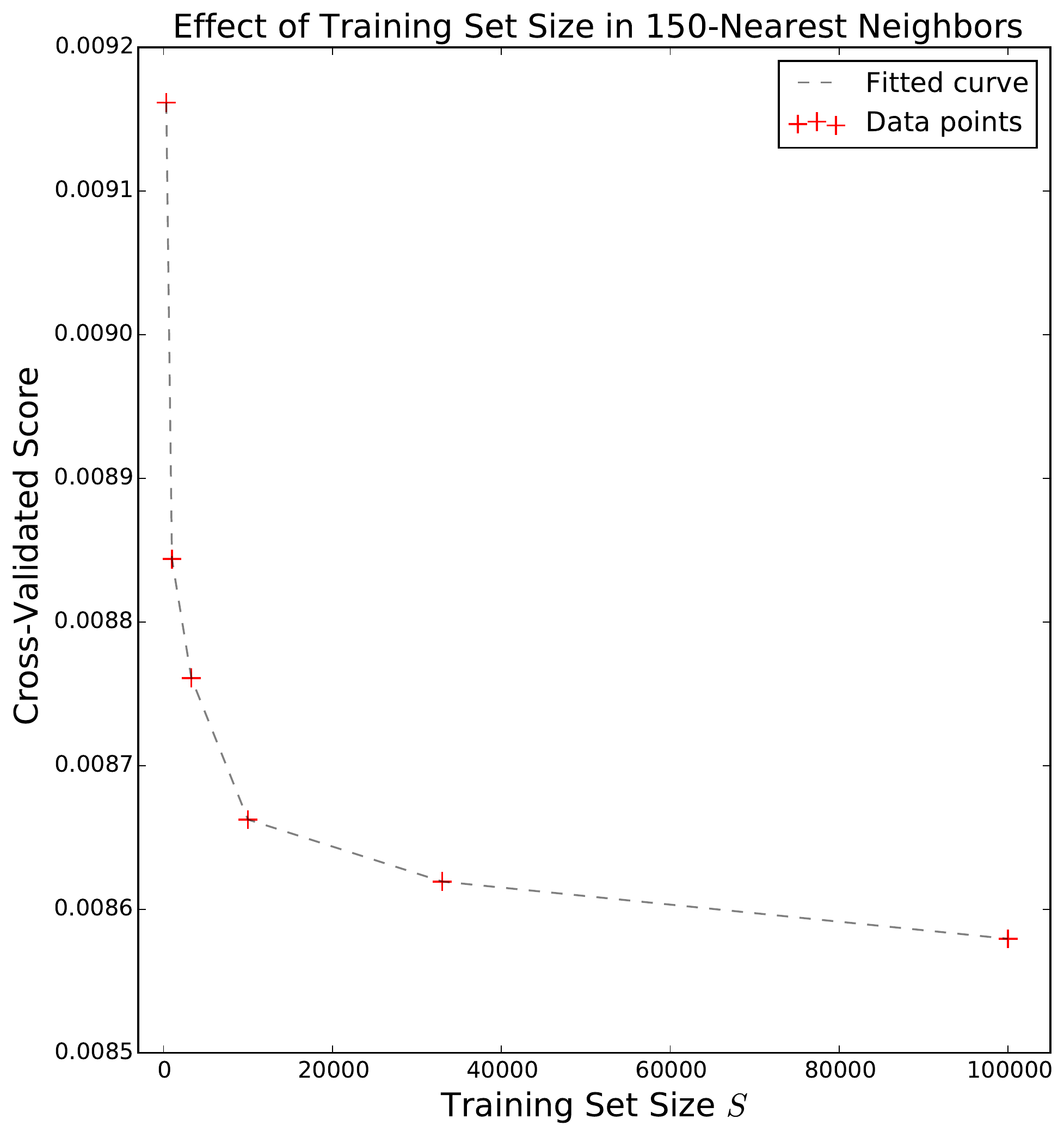}
	       \caption{Effect of Training Set Size $S$ on 150-$n$ Neighbors}
		   \label{fig:scoresVsS}
	   \end{figure}

	   Once a nearly optimal $k$ was identified, we measured the effect that increasing the training set size had
	   on classifier score on a 100,000 point validation set for this particular $k$.
	   We searched by factors of $\sqrt(10)$ from 3.33e2 to 1.00e5 and found 
	   performance increased monotonically with larger training set sizes.
	   One salient feature of this graph is that large initial gains in performance
	   eventually gave way to much smaller later gains, so that 
	   the model seems to have some a threshold effect as data size increases.
	   See Figure~\ref{fig:scoresVsS}.
	   In particular, we observed less than a unit increase in score performance
	   for a unit log increase in training set size.
	   One possible explanation for this is that the classifier is approaching the Bayes error.
	   Predictions on the test set were carried out using the 100,000 point training set
	   on the far right end of the graph. 
	    
	\subsection{Performance}
	    For our near optimal $k$ of 150 and training set size of 100,000,
		we obtained a validation set score of 0.00857948.
		A score of 0.00857948 is much better than all the other algorithms tested,
		would be good enough for 73 world-wide out of 331 in the Kaggle competition,
		placing it in the top 25\% of submissions world-wide. 
		
		Further improvements are expected to give more gains.
		First among these is that we have a training set size of ~1 million to build a bigger, better k-d tree,
		though this effect is expected to be limited as noted above.
		Another approach is parameter optimization, in particular $p$, the
		$l_p$ notion of distance used in the nearest neighbor search.
		A further avenue towards improvement is filling in missing data 
		in some more intelligent way than as was done simply with zeros,
		for instance with either column means or medians.
		We hope to pursue all of these avenues. 
	
\section{Summary of Results}
	
\begin{figure}
        \includegraphics[width=0.95\textwidth]{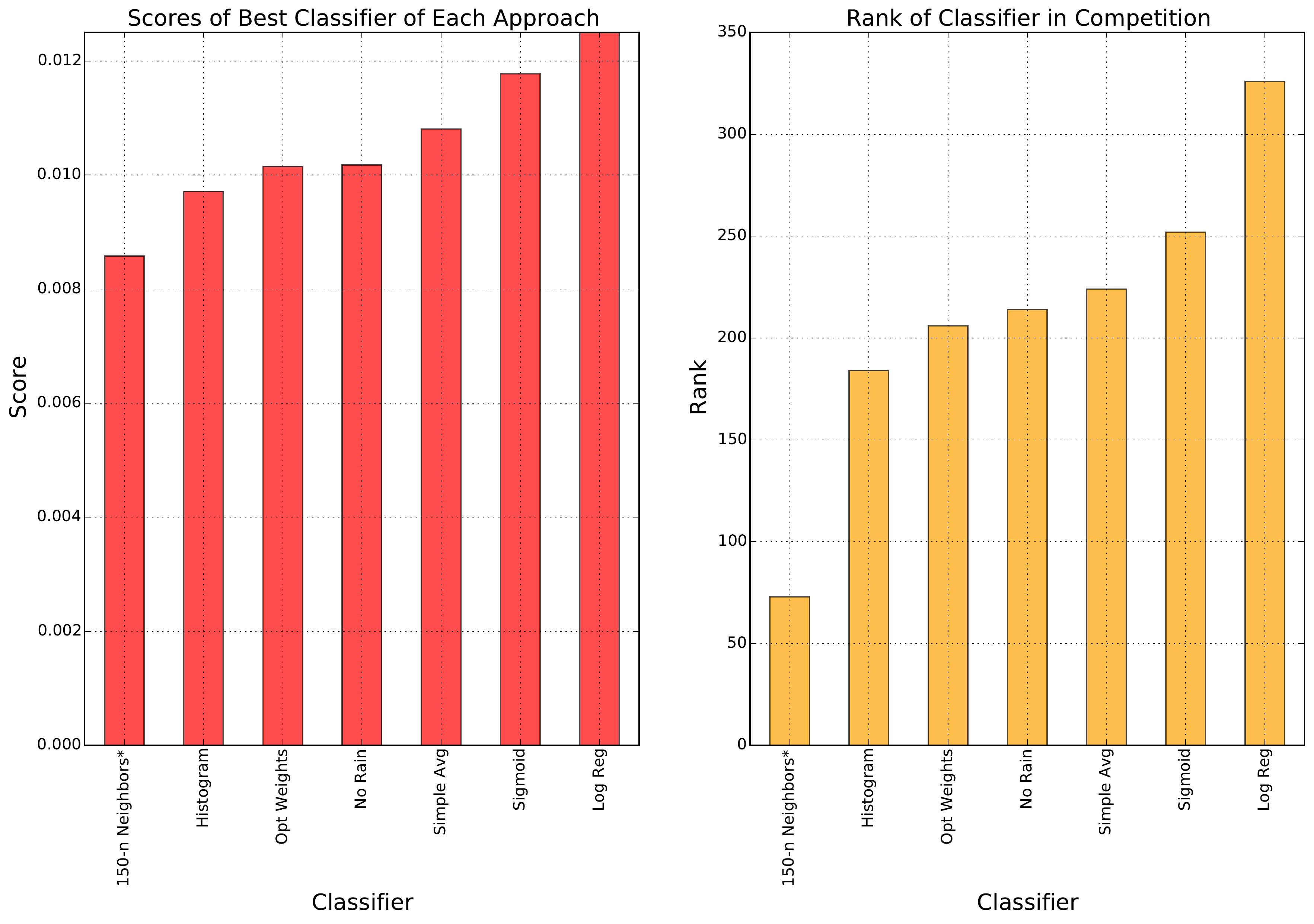}
		\caption{Summary of scores and ranking for classifiers}
		\label{fig:summary}
	\end{figure}

	Figure~\ref{fig:summary} summarizes the scores and ranking for the various classifiers tried.
	Nearest neighbors was the best classifier tested, followed by `Histogram'.
	
\section{Conclusions}
    \subsection{Further Directions}
		\subsubsection{More Derived Features}
		   An immediate area for further exploration is a richer set of derived features from the given time series
		   features. Only series averages were computed as derived features, and
		   this is suspected as one bottleneck in predictor performance. 
        \subsubsection{LDA, QDA}
		   Linear Discriminant Analysis (LDA) and Quadratic Discriminant Analysis (QDA)
		   are two further parametric models to explore,
		   with QDA providing greater decision boundary flexibility.
		   Both approaches however assume more than Logistic Regression models do, 
		   in particular a Gaussian distributions of labels,
		   and suffer in performance to varying degrees when this assumption does not hold.
	    \subsubsection{Other ML Methods, SVM's, SVR's, Neural Networks}
		   A whole slew of other machine learning approaches could be applied to these data.
		   Support vector (machine) regressors, or SVR's, stand out, 
		   given their computational efficiency and other favorable theoretical features.
		   Neural networks would be intriguing to explore, in particular their ability 
		   for automatic feature detection. 
		   An approach for using classifiers like SVM's in this problem to produce probability
		   distributions is sketched below.
	    \subsubsection{From Classifiers to Probability Distribution Estimators}
		    One standard way to turn classifiers into regressors is to create many classification labels. 
			Another idea for this problem is to 
			use classifiers to update some prior probability density estimates: 
			First we take some posterior probability density estimates drawn from the training histogram
			or from a nearest neighbor search.
			Then we update the estimate based on what the classifier predicts.
			The strength of the update can be take as a tunable parameter optimized through cross-validation.
			One concrete way to think about this update would be that the posterior probability density estimate
			is 100 votes about which bin the test sample is at.
			If a classifier predicts the test sample is in bin $x$, then
			add $r$ many votes for bin $x$.
		\subsubsection{Ensemble of Best Performing Models}
		    One last approach is always to take an ensemble of all the best performing
			classifiers from each of the above approaches.
	    
\section*{Acknowledgements}
    Thank you to Professor Bartlett and Professor Efros for very useful and insightful discussions 
	all the way from thinking about ideas for the project through working out particular algorithms for this data set.
	I felt like I learned so much in the course and through this project, and very much appreciated your help!


\end{document}